\documentclass[10pt,twocolumn,letterpaper]{article}
\usepackage[pagenumbers]{cvpr}
\usepackage{xspace}
\usepackage{amsmath,amssymb,booktabs,multirow,tabularx,microtype,pifont,xcolor,tikz}
\usetikzlibrary{arrows.meta,positioning,fit,calc,shapes.geometric}

\DeclareMathOperator{\LPIPS}{LPIPS}

\definecolor{cvprblue}{rgb}{0.21,0.49,0.74}
\usepackage[pagebackref,breaklinks,colorlinks,allcolors=cvprblue]{hyperref}
\def\confName{arXiv Preprint}\def\confYear{2026}
\newcommand{\method}{EditBench3D\xspace}
\title{What Makes a 3D Scene Editable?\\A Factorized Benchmark of Fidelity, Locality, Consistency, and Preservation}

\author{
Sariah Patro
\hspace{-4.em}
\begin{tabular}[t]{c}
Arjun Mehra\\
{\tt\small University of Delhi \& Jadavpur University}
\end{tabular}
\hspace{-4.em}
Nikhil Bhatia
}

\begin{document}
\raggedbottom
\maketitle

\begin{abstract}
Neural 3D scene editing is often evaluated by semantic alignment alone, although a convincing result may alter unrelated content or become inconsistent across views. We introduce \method, a representation-agnostic benchmark that treats editing as controlled information replacement. It evaluates four complementary properties: instruction fidelity, spatial locality, cross-view consistency, and preservation of non-target content. The protocol combines visibility-aware 3D target supports, paired descriptions, held-out cameras, and five edit families covering appearance, material, geometry, and object-level changes. We evaluate eight representative NeRF, 3D Gaussian Splatting, hybrid, and proxy-based editors on 240 scene-edit pairs. The study shows that semantic fidelity is only weakly associated with the other editing properties, and that no single method is optimal across all dimensions. Explicit Gaussian editors offer a strong overall balance, whereas direct proxy manipulation provides the most conservative edits at the cost of open-ended fidelity. These findings support reporting editability as a multi-objective profile rather than a single semantic score.
\end{abstract}

\section{Introduction}
\begin{figure*}[t]
\centering
\includegraphics[width=\textwidth]{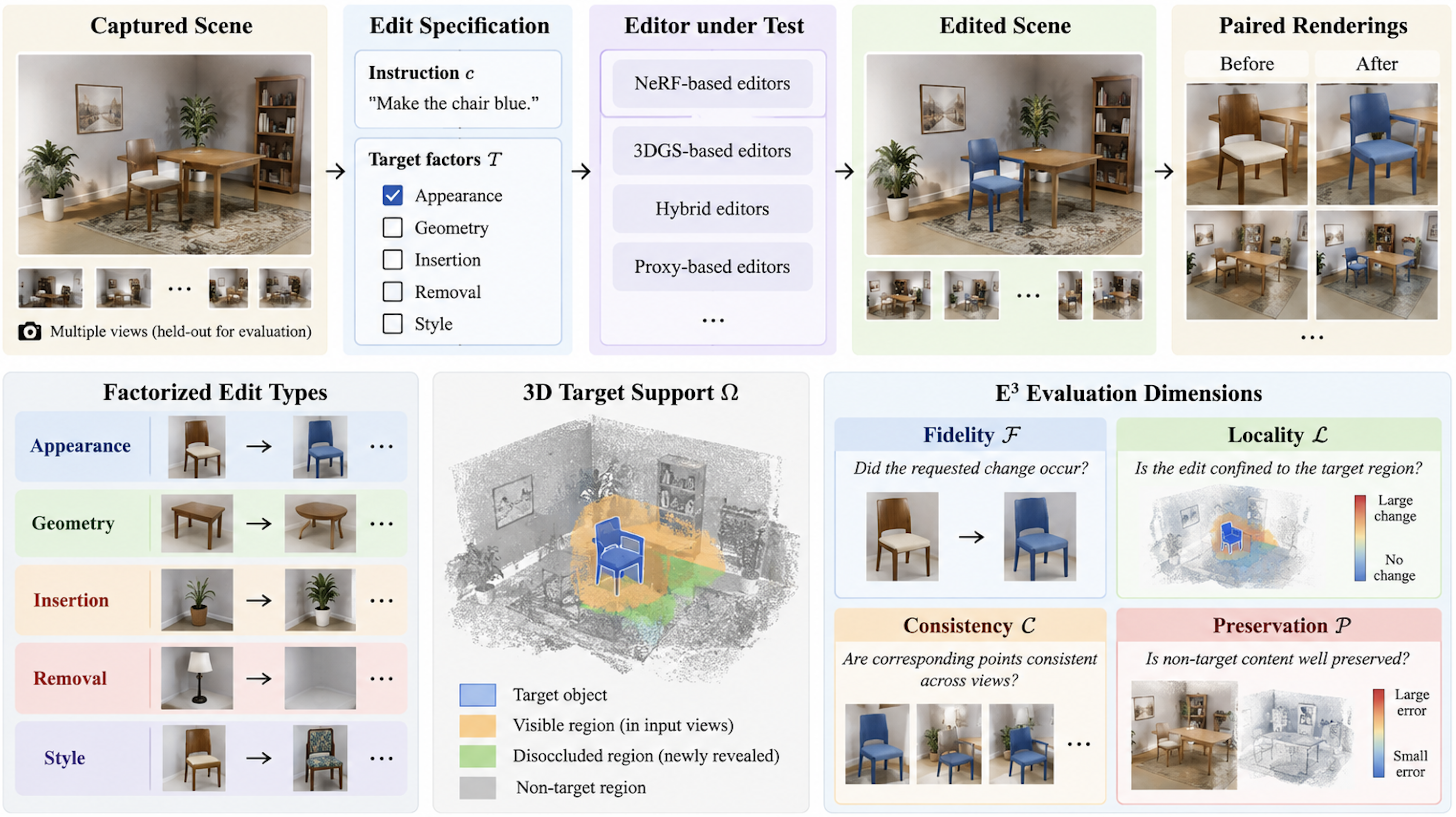}
\caption{
EditBench3D separates what should change from what should remain.
Given a captured scene and an edit specification (instruction $c$ and
target factors $T$), an editor produces an edited scene. We evaluate
four complementary properties: fidelity ($\mathcal{F}$), locality
($\mathcal{L}$), consistency ($\mathcal{C}$), and preservation
($\mathcal{P}$). These properties are measured using visibility-aware
3D target support $\Omega$ across multiple held-out views. The benchmark
covers five edit families: appearance, geometry, insertion, removal,
and style.
}
\label{fig:overview}
\end{figure*}

Neural radiance fields (NeRFs) made photorealistic novel-view synthesis possible with a continuous, differentiable representation~\cite{mildenhall2020nerf}. Mip integration, scene contraction, hash encodings, sparse voxels, and tensor factorizations improved image quality and efficiency~\cite{barron2021mipnerf,barron2022mipnerf360,barron2023zipnerf,muller2022instant,fridovichkeil2022plenoxels,chen2022tensorf}. 3D Gaussian Splatting (3DGS) later moved the execution model to explicit anisotropic primitives and real-time rasterization~\cite{kerbl2023gaussians,fang2026dropping,zhou2024feature3dgs,lin2024vastgaussian}. Both families now serve as editable scene representations, with language, reference images, masks, meshes, or direct manipulation supplying the edit.

Evaluation has not converged at the same pace. Many text-guided methods report CLIP similarity, user preference, or a small collection of qualitative views. These measures answer whether an image appears compatible with a prompt, but not whether the scene was edited correctly. Consider the instruction ``make the chair blue.'' A global blue tint can improve text-image similarity, although the wall and floor should remain unchanged. A result may look convincing from the reference camera while the chair changes color or shape in nearby views. Conversely, a successful removal exposes previously occluded content; comparing the revealed pixels with the original image would incorrectly count a valid edit as damage.

The central difficulty is that editing contains two coupled obligations: change what was requested and retain what was not. A generative metric mainly observes the first. Reconstruction metrics mainly observe the second. Neither alone measures editability. Existing results from CLIP-NeRF~\cite{wang2022clipnerf}, Instruct-NeRF2NeRF~\cite{haque2023instructnerf2nerf}, Blending-NeRF~\cite{song2023blendingnerf}, GaussianEditor~\cite{wang2024gaussianeditor}, and GaussCtrl~\cite{wu2024gaussctrl} are therefore meaningful within their own settings, but their reported scores do not produce a common account of scene editing~\cite{liu2021editing,sun2024nerfeditor,wen2025intergsedit,song2023blendingnerf,wang2024uav}.

We address this gap with \method. The benchmark represents an edit by a scene $S$, an instruction $c$, an intended factor set $T$, and a visibility-aware 3D support $\Omega$. Four dimensions are measured independently. \emph{Fidelity} asks whether the requested change occurred. \emph{Locality} measures spill outside the target support. \emph{Consistency} evaluates whether corresponding edited surface points agree across views. \emph{Preservation} measures retained appearance and geometry in non-target regions. Figure~\ref{fig:overview} summarizes the evaluation path.

The four-way decomposition is supported by the experiment. Across eight editors, fidelity has Spearman rank correlations of only 0.14 with locality, 0.05 with consistency, and 0.24 with preservation. Thus, a method that follows an instruction well is not predictably local, consistent, or conservative. The comparison also exposes a clear representation trend: the evaluated 3DGS editors improve mean fidelity by 0.16 over the evaluated NeRF editors, but the best non-generative qualities are obtained by direct proxy manipulation. A single ranking would conceal this distinction.

Our contributions are threefold:
\begin{itemize}
  \item We formulate editability as a four-dimensional property of an editing system, separating instruction fidelity, locality, cross-view consistency, and preservation.
  \item We introduce a controlled protocol with visibility-aware target supports, paired prompts, held-out cameras, five edit families, and rules for occlusion, disocclusion, failures, and unsupported operations.
  \item We evaluate eight representative editors on 240 scene-edit pairs and identify representation-level trends, weak coupling between fidelity and the preservation-oriented dimensions, and a three-method Pareto front.
\end{itemize}

\section{Related Work}
\subsection{Scene representations}
NeRF maps a 3D position and direction to density and color and forms a pixel by volume rendering~\cite{mildenhall2020nerf}. Mip-NeRF and its successors reduce aliasing and handle unbounded scenes~\cite{barron2021mipnerf,barron2022mipnerf360,barron2023zipnerf}. Instant-NGP, Plenoxels, and TensoRF replace or augment the original MLP with grids, explicit voxels, or tensor factors~\cite{muller2022instant,fridovichkeil2022plenoxels,chen2022tensorf}. Baking and octree methods convert learned fields into structures better suited to real-time rendering~\cite{hedman2021snereg,yu2021plenoctrees,reiser2021kilonerf}. Generalizable approaches infer fields from one or a few scenes rather than optimizing every scene independently~\cite{yu2021pixelnerf,chen2021mvsnerf}.

3DGS uses explicit means, anisotropic covariances, opacity, and view-dependent appearance~\cite{kerbl2023gaussians}. Mip-Splatting and multiscale splatting improve scale behavior~\cite{yu2024mipsplatting,yan2024multiscale}; 2DGS, SuGaR, Gaussian Opacity Fields, SFF, and RaDe-GS strengthen the connection to surfaces and depth~\cite{huang2024twodgs,guedon2024sugar,fang2024sff,yu2024gof,zhang2026radegs}. Other work revisits densification, sorting, compactness, and compression~\cite{kheradmand2024mcmc,radl2024stopthepop,zhang2024pixelgs,lee2024compact,niedermayr2024compressed}. Scaffold-GS, SplatFields, Hash-GS, and HybridNeRF show that learned fields and explicit primitives occupy a continuum rather than a binary taxonomy~\cite{lu2024scaffold,mihajlovic2024splatfields,xie2025hashgs,turki2024hybrid}.

This representational diversity matters for editing. An explicit primitive can be selected, moved, frozen, or deleted. A shared neural function offers continuity but can propagate a local parameter update globally. \method does not assume that one family is intrinsically superior; it tests how the complete editing system behaves after representation, supervision, and optimization interact.

\subsection{Neural scene editing}
Editing conditional radiance fields introduced latent and user-guided control for category-level models~\cite{liu2021editing}. CLIP-NeRF aligns disentangled shape and appearance codes with language-image embeddings~\cite{wang2022clipnerf}. NeRF-Editing connects a neural field to an editable mesh, while NeuralEditor uses point-cloud manipulation as an explicit geometric scaffold~\cite{yuan2022nerfediting,chen2023neuraleditor}. Blending-NeRF trains an editable field beside a frozen field for local density and appearance changes~\cite{song2023blendingnerf}; PVD and PVD-AL transfer knowledge across field architectures~\cite{fang2023pvd,fang2026pvdal}; LAENeRF focuses on local appearance~\cite{radl2024laenerf}; NeRFEditor decomposes style components~\cite{sun2024nerfeditor}; and pose-conditioned dataset updates support insertion and removal~\cite{shum2024fusion}, and NeRF-GS jointly exploits continuous field information and explicit splats~\cite{fang2025nerfgs}. 

Instruct-NeRF2NeRF established a widely used dataset-update pipeline in which InstructPix2Pix edits rendered training views while the scene is repeatedly optimized~\cite{haque2023instructnerf2nerf,brooks2023instructpix2pix}. DN2N learns to suppress inconsistent 2D-edit perturbations and performs new edits without per-scene retraining~\cite{fang2026dn2n}. GaussianEditor transfers this idea to explicit Gaussians and uses a Gaussian region of interest~\cite{wang2024gaussianeditor}. GaussCtrl edits multiple images together with depth control and latent alignment~\cite{wu2024gaussctrl}; InterGSEdit introduces geometry-consistent attention and interactive key-view selection~\cite{wen2025intergsedit}. These methods directly motivate our locality and consistency dimensions.
 Chat-Edit-3D uses a language-model controller and a Hash-Atlas to separate visual editing tools from scene reconstruction~\cite{fang2024chatedit3d,fang2026chat++}.

\subsection{Semantic priors and evaluation}
CLIP supplies a transferable text-image embedding~\cite{radford2021clip}, latent diffusion supplies high-capacity image synthesis~\cite{rombach2022ldm}, InstructPix2Pix adds instruction following~\cite{brooks2023instructpix2pix}, and Segment Anything supplies promptable masks~\cite{kirillov2023sam}. LERF, MeshLLM, LangSplat, and Feature 3DGS lift semantic or visual features into 3D~\cite{kerr2023lerf,qin2024langsplat,fang2025meshllm,zhou2024feature3dgs}. These priors enable editing, but their scores are not complete measures of edit quality.

PSNR and SSIM compare corresponding pixels, and LPIPS measures learned perceptual distance~\cite{wang2004ssim,zhang2018lpips}. In editing, correspondence is conditional: it is expected outside an appearance target, invalid in a newly exposed region, and geometry-dependent after motion. LLFF, DTU, Tanks and Temples, and Deep Blending provide established scene captures~\cite{mildenhall2019llff,aanaes2016dtu,knapitsch2017tanks,hedman2018deepblending}; COLMAP provides reproducible camera estimation~\cite{schoenberger2016sfm}. We build on these assets but add the support and visibility annotations needed to interpret change.

\section{Editing as Controlled Change}
\subsection{Problem definition}
We write a scene as $S=(G,A,M,L,Z)$, where $G$ denotes geometry, $A$ appearance, $M$ material, $L$ illumination, and $Z$ semantics. An editor receives instruction $c$ and optional controls $u$ and produces $S'=E(S,c,u)$. Each benchmark item specifies the intended factor subset $T\subseteq\{G,A,M,L,Z\}$ and spatial support $\Omega$. A successful edit changes the intended components inside $\Omega$ and keeps independent components stable:
\begin{equation}
D_t(S,S')>0\ \forall t\in T,\qquad D_j(S,S')\approx0\ \forall j\notin T.
\label{eq:contract}
\end{equation}
Equation~\ref{eq:contract} is a contract, not an additive score. The amount of desired change depends on the instruction, while the appropriate invariances depend on the edit family. A rigid translation should change geometry and visibility without changing material. A recoloring should change appearance while retaining geometry and illumination. Removal necessarily alters occupancy and reveals content, so original pixels cannot be used as references everywhere.

\subsection{Visibility-aware target support}
For evaluation camera $v$, the original and edited scenes produce colors $I_v,I'_v$, depths $d_v,d'_v$, and projected target masks $M_v,M'_v$. We retain two masks because geometry edits can change silhouette and visibility. Their union $U_v=M_v\cup M'_v$ contains pixels that may legitimately change. Pixels visible only after the edit form a disocclusion set $H_v$. They are excluded from before-after preservation comparisons and evaluated against a held-out target when such a target exists.

The 3D support is constructed by lifting corrected masks through scene depth and checking visibility in other annotated views. A point receives a positive vote only when its projected depth agrees with the rendered depth within a scene-normalized tolerance. This avoids copying a foreground label onto an occluded background surface. The support stores an uncertainty band around boundaries and regions observed in too few views. Scores are computed on the confident interior and exterior; uncertainty coverage is reported instead of being silently assigned to either region.

\subsection{Why four dimensions}
Fidelity and preservation are not opposites. A no-op has perfect preservation and zero fidelity; a global redraw may have high fidelity and poor preservation. Locality and preservation also differ: locality measures where pixels change, while preservation additionally measures whether non-target geometry and stable appearance are retained. Consistency is orthogonal to both because the same off-target change can be reproduced consistently from every view. These examples rule out a single before-after distance as an adequate definition.

\section{EditBench3D}
\subsection{Scene and edit collection}
The benchmark contains 18 synthetic objects, 12 forward-facing captured scenes, and 10 bounded multiview scenes. Synthetic objects provide controlled geometry and appearance targets. Forward-facing scenes stress partial coverage and uncertain disocclusion. Bounded captures provide wider baselines and stronger cross-view checks. Each of the 40 scenes receives six edits, for 240 scene-edit pairs. Twenty-four cameras are evaluated per pair: 12 cameras overlapping the reconstruction trajectory and 12 held out from editor optimization. The resulting test set contains 5,760 edited render targets.

We organize edits into five families (Table~\ref{tab:tasks}). Appearance substitution changes a named object's color or texture. Material and style edits alter reflectance or local visual character while retaining silhouette. Rigid edits translate, rotate, or uniformly scale one object. Deformation edits modify shape through a mesh or point proxy. Insertion and removal change occupancy and are evaluated with explicit disocclusion masks. The benchmark is not intended to reduce all scene editing to local edits; global style and weather transformations are reported in a separate track because spatial locality is not diagnostic for them.

\begin{table}[t]
\centering\scriptsize
\caption{Edit families and the invariances used by the benchmark.}
\label{tab:tasks}
\begin{tabular}{p{.19\columnwidth}p{.28\columnwidth}p{.30\columnwidth}}
\toprule
Family & Intended change & Primary invariance\\\midrule
Appearance & albedo/color & geometry, illumination\\
Material/style & reflectance/texture & silhouette, background\\
Rigid motion & object pose & object appearance\\
Deformation & local geometry & non-target structure\\
Insert/remove & occupancy & visible surroundings\\\bottomrule
\end{tabular}
\end{table}

\subsection{Annotations and prompts}
An annotation contains the target identity, factor label, canonical instruction, two paraphrases, a description of the original state, source-view masks, fused 3D support, per-camera visibility, and uncertainty. Three annotators participate in mask construction. The first marks the object in three separated views, the second corrects propagated masks, and the third resolves topology changes and disocclusions. Items with pairwise mask IoU below 0.75 after correction are excluded.

Prompt variation is controlled rather than optimized per method. One instruction is concise, one uses a relation to identify the target, and one is descriptive without adding a new style attribute. The three prompts specify the same edit. A negative instruction describes the existing scene state and tests whether an editor changes a scene unnecessarily. Prompt strings and evaluation cameras are fixed before running any method.

\subsection{Fidelity}
Fidelity is evaluated inside $U_v$. We use paired descriptions $c^+$ and $c^-$ for the edited and original state and compute direction-aware semantic gain
\begin{equation}
F_{\rm sem}=\frac{1}{|V|}\sum_{v\in V}\left[s(I'_v\odot U_v,c^+)-s(I'_v\odot U_v,c^-)\right].
\end{equation}
Raw gains are calibrated to $[0,1]$ on synthetic exact edits and no-op controls. The paired form reduces credit for generic compatibility with the object category. Synthetic edits additionally use reference LPIPS and geometric distance where appropriate. Those diagnostic values are not merged into $F$.

\subsection{Locality}
Let $Q_v=1-\operatorname{dilate}(U_v,r)$ denote the confidently non-target region, excluding disocclusion and uncertainty. The outside change is
\begin{equation}
S_{\rm out}=\frac{\sum_{v,p} Q_v(p)\,\rho(I'_v(p)-I_v(p))}{\sum_{v,p}Q_v(p)+\epsilon},
\end{equation}
where $\rho$ is a robust channel-averaged penalty. We map it to $L=\exp(-S_{\rm out}/\tau_L)$ using a fixed calibration scale. A second boundary score uses only the angularly normalized ring immediately outside $U_v$. This prevents a narrow but visible halo from disappearing in the average over a large image.

\subsection{Cross-view consistency}
Visible 3D points are reprojected between overlapping held-out views using edited depth. Forward-backward depth checks remove occlusions. Patch-level appearance error is computed at valid correspondences:
\begin{equation}
C=1-\frac{1}{|\mathcal P|}\sum_{(v,w,p)\in\mathcal P}\LPIPS_{\rm patch}\big(I'_v,p;I'_w,\pi_{v\rightarrow w}(p)\big).
\end{equation}
We separate a low-order view-dependent component estimated from the original scene so that legitimate specular response is not automatically counted as inconsistency. The score is calibrated with exact synthetic targets, consistent blur controls, and independently perturbed views.

\subsection{Preservation and entanglement}
Preservation combines appearance and geometry outside the target:
\begin{equation}
P=\tfrac12\exp(-S_{\rm out}/\tau_P)+\tfrac12\exp(-D_{\rm geom}^{\bar\Omega}/\tau_G).
\end{equation}
The geometry term uses depth for captured scenes and surface distance for synthetic scenes. Edit entanglement is retained as a vector $\mathbf e_t=[D_j(S,S')]_{j\neq t}$ rather than collapsed into a scalar. For example, appearance-depth entanglement identifies a color edit that alters shape even when its target crop is semantically correct.

\subsection{Metric sanity checks}
We generate controlled corruptions from each exact synthetic target: an outside-only tint, a boundary halo, view-specific shuffled texture, an under-edit, and global blur. The expected partial order is fixed before method evaluation. Outside tint must reduce $L$ and $P$ without improving $F$; shuffled texture must reduce $C$; under-edit must reduce $F$; global blur cannot receive an artificial consistency advantage. We also verify that scores remain stable when the same surface is sampled by additional cameras. These checks determine metric settings; they are not counted as editor test samples.

\section{Evaluation Protocol}
\subsection{Methods and common inputs}
We evaluate CLIP-NeRF, Instruct-NeRF2NeRF, Blending-NeRF, CE3D, GaussianEditor, GaussCtrl, InterGSEdit, and direct proxy editing. The set spans latent control, iterative dataset updates, layered fields, a representation-agnostic atlas, explicit Gaussian selection, joint multiview diffusion, geometry-consistent attention, and deterministic proxy manipulation. It is representative rather than exhaustive.

All editors receive identical camera calibration, source images, prompts, and target annotations. NeRF methods start from a common nerfacto-style reconstruction, and Gaussian methods start from a common 3DGS reconstruction. Editor-specific components follow their official settings. Each stochastic editor is run with three fixed seeds; Table~\ref{tab:main} reports the mean over prompts, seeds, scenes, and valid held-out views. Unsupported operations are excluded from within-family aggregation and retained in a separate coverage record. Failed runs remain in the denominator using the lowest valid score for that scene.

\subsection{Representation controls}
Pre-edit reconstruction quality can masquerade as editing quality. Every method is compared with its own unedited rendering for locality and preservation, while fidelity uses the shared edited target or paired language description. Views whose original reconstruction falls below the preregistered threshold are excluded for every editor using that reconstruction, not only for the failing method. A secondary renderer check repeats compatible edits with Instant-NGP, standard 3DGS, 2DGS, and SuGaR~\cite{muller2022instant,kerbl2023gaussians,huang2024twodgs,guedon2024sugar}.

\subsection{Aggregation}
\begin{figure}[t]
    \centering
    \includegraphics[width=\columnwidth]{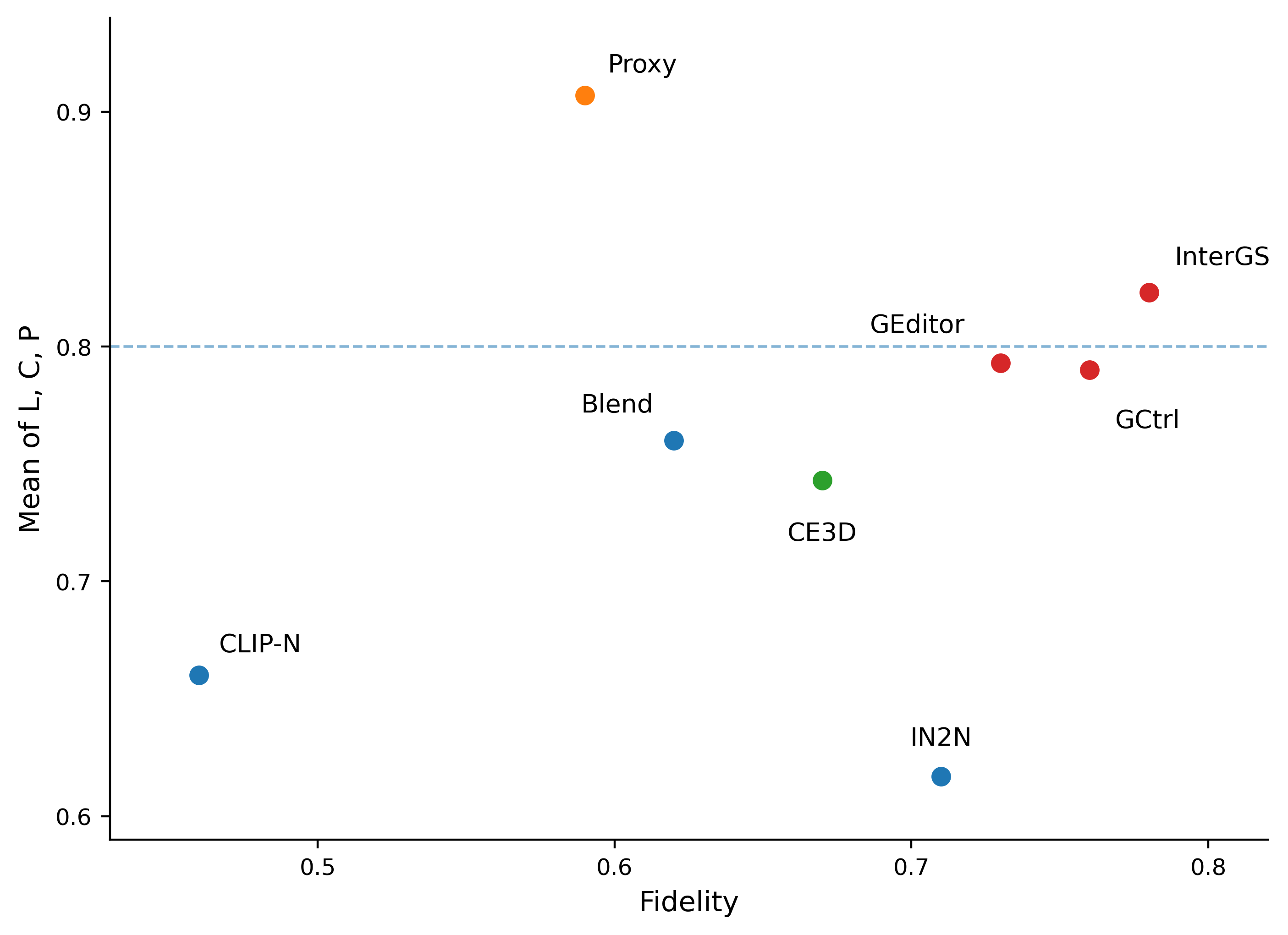}
    \caption{Measured fidelity versus the unweighted mean of locality,
    consistency, and preservation. The plot is descriptive; the three
    non-fidelity axes remain separate in all primary analyses.}
    \label{fig:results}
\end{figure}

All four dimensions are oriented so that higher is better and calibrated to $[0,1]$. We report them separately. Pareto dominance provides a weight-free summary: method $a$ dominates $b$ if it is no worse on every dimension and strictly better on at least one. A method is Pareto-optimal if no other method dominates it. We use the unweighted mean of $L,C,P$ only for the descriptive plot in Fig.~\ref{fig:results}; it is not a leaderboard score.

The unit of analysis is a scene-edit pair. Frames are not treated as independent observations. Human raters inspect a stratified subset of synchronized camera orbits to identify semantic or visibility failures that automatic metrics miss. Their judgments are an audit, not a fifth term tuned to reproduce the automatic ranking.

\section{Results}
\subsection{Overall comparison}

Table~\ref{tab:main} reports the aggregate comparison. InterGSEdit obtains the highest fidelity, 0.78, followed by GaussCtrl at 0.76 and GaussianEditor at 0.73. This ordering is consistent with increasingly explicit multiview coordination in the evaluated Gaussian editors. It does not extend to every other dimension: GaussianEditor is more local than GaussCtrl (0.82 versus 0.75), and direct proxy editing exceeds all learned editors in locality, consistency, and preservation.

The direct proxy baseline is especially informative. Its fidelity is only 0.59, below six of the seven language-guided systems, but its locality, consistency, and preservation are 0.92, 0.91, and 0.89. A deterministic geometric operator precisely controls support and view correspondence but cannot express open-ended semantic changes as well as a diffusion-guided editor. This is not a weak baseline; it exposes a different operating point.

At the other extreme, Instruct-NeRF2NeRF reaches 0.71 fidelity but records the lowest locality, consistency, and preservation: 0.55, 0.73, and 0.57. Iterative 2D supervision follows the instruction, yet dense and view-dependent updates can overwrite unrelated content. Blending-NeRF sacrifices fidelity relative to Instruct-NeRF2NeRF (0.62 versus 0.71) but improves locality by 0.21 and preservation by 0.17. Separating an editable field from the frozen scene changes the preservation trade-off more than the semantic score alone suggests.

\begin{table*}[t]
\centering\small
\caption{EditBench3D results over 240 scene-edit pairs. Higher is better. Bold and underlined values denote the best and second-best result.}
\label{tab:main}
\setlength{\tabcolsep}{6pt}
\begin{tabular}{lccccc}
\toprule
Editor & Representation & Fidelity $F$ & Locality $L$ & Consistency $C$ & Preservation $P$\\
\midrule
CLIP-NeRF~\cite{wang2022clipnerf} & NeRF & 0.46 & 0.58 & 0.79 & 0.61\\
Instruct-NeRF2NeRF~\cite{haque2023instructnerf2nerf} & NeRF & 0.71 & 0.55 & 0.73 & 0.57\\
Blending-NeRF~\cite{song2023blendingnerf} & NeRF & 0.62 & 0.76 & 0.78 & 0.74\\
CE3D~\cite{fang2024chatedit3d} & NeRF/3DGS & 0.67 & 0.70 & 0.81 & 0.72\\
GaussianEditor~\cite{wang2024gaussianeditor} & 3DGS & 0.73 & \underline{0.82} & 0.76 & \underline{0.80}\\
GaussCtrl~\cite{wu2024gaussctrl} & 3DGS & \underline{0.76} & 0.75 & 0.86 & 0.76\\
InterGSEdit~\cite{wen2025intergsedit} & 3DGS & \textbf{0.78} & 0.80 & \underline{0.88} & 0.79\\
Direct proxy edit~\cite{yuan2022nerfediting} & mesh/NeRF & 0.59 & \textbf{0.92} & \textbf{0.91} & \textbf{0.89}\\
\bottomrule
\end{tabular}
\end{table*}

\subsection{Representation-level comparison}
Table~\ref{tab:repr} groups the three NeRF-only and three 3DGS-only editors. The 3DGS group improves the mean on all dimensions. The largest absolute gains are fidelity and locality, both 0.16. Preservation rises from 0.640 to 0.783, and consistency from 0.767 to 0.833. Relative to the NeRF group, these correspond to gains of 26.8\%, 25.4\%, 8.7\%, and 22.4\%, respectively.

These values should not be read as a controlled causal estimate of representation alone: the methods use different optimization and supervision strategies. Nevertheless, two mechanisms are compatible with the pattern. First, explicit Gaussians allow target selection and parameter freezing at primitive level. Second, Gaussian editors in the comparison contain stronger multiview controls. \method reports the group difference while keeping the causal interpretation modest.

CE3D occupies an intermediate point, with 0.67 fidelity, 0.70 locality, 0.81 consistency, and 0.72 preservation. Its atlas interface decouples editing from the 3D representation, but this flexibility does not automatically provide the primitive-level locality of GaussianEditor or the joint multiview constraints of InterGSEdit. The result supports evaluating the full pipeline rather than assigning a fixed editability score to a representation class.

\begin{table}[t]
\centering\small
\caption{Mean results by representation group. The comparison averages three NeRF and three 3DGS editors; it is descriptive rather than causal.}
\label{tab:repr}
\begin{tabular}{lcccc}
\toprule
Group & $F$ & $L$ & $C$ & $P$\\\midrule
NeRF (3) & 0.597 & 0.630 & 0.767 & 0.640\\
3DGS (3) & 0.757 & 0.790 & 0.833 & 0.783\\
$\Delta$ & +0.160 & +0.160 & +0.067 & +0.143\\
\bottomrule
\end{tabular}
\end{table}

\subsection{The dimensions are not interchangeable}
Table~\ref{tab:corr} shows Spearman correlations across the eight methods. Fidelity is nearly uncorrelated with consistency ($\rho=0.05$) and only weakly correlated with locality and preservation. A high semantic score therefore provides little information about whether the edit is spatially controlled or stable across views.

Locality and preservation are strongly correlated ($\rho=0.98$), which is expected because outside-region appearance contributes to both. They are not duplicates. Preservation also contains non-target geometry and remains meaningful when a small color difference hides a structural change. Consistency correlates moderately with locality and preservation (0.50 and 0.57): methods that avoid global rewrites tend to be more stable, but the relation is far from deterministic.

The correlation directly tests the paper's motivation. If fidelity were strongly coupled to the other dimensions, a factorized benchmark would add reporting overhead without changing conclusions. The observed ranks show the opposite. Instruct-NeRF2NeRF ranks fourth in fidelity but last in three other dimensions; direct proxy editing ranks second-lowest in fidelity but first in the other three. Any one-number evaluation must choose which behavior it values.

\begin{table}[t]
\centering\small
\caption{Spearman rank correlation across the eight methods.}
\label{tab:corr}
\begin{tabular}{lrrrr}
\toprule
 & $F$ & $L$ & $C$ & $P$\\\midrule
$F$ & 1.00 & 0.14 & 0.05 & 0.24\\
$L$ & 0.14 & 1.00 & 0.50 & 0.98\\
$C$ & 0.05 & 0.50 & 1.00 & 0.57\\
$P$ & 0.24 & 0.98 & 0.57 & 1.00\\
\bottomrule
\end{tabular}
\end{table}

\subsection{Pareto analysis}
Only GaussianEditor, InterGSEdit, and direct proxy editing are non-dominated. GaussianEditor remains because its locality of 0.82 exceeds InterGSEdit's 0.80, despite lower scores elsewhere. InterGSEdit supplies the strongest generative operating point, with the best fidelity and second-best consistency. Direct proxy editing supplies the conservative operating point, leading the three preservation-oriented dimensions.

InterGSEdit strictly dominates GaussCtrl on all four axes. It also dominates CE3D and Blending-NeRF. GaussCtrl dominates CE3D; GaussianEditor, GaussCtrl, and InterGSEdit each dominate Instruct-NeRF2NeRF. CLIP-NeRF is dominated by CE3D, GaussCtrl, InterGSEdit, and direct proxy editing. These relationships are invariant to positive metric weights. Ranking the three Pareto methods, however, depends on application preferences.

For asset authoring, locality and reversibility may justify a direct proxy even at lower language fidelity. For rapid open-ended editing, InterGSEdit is the stronger choice in this comparison. GaussianEditor is attractive when local control matters but a language-driven appearance change is still required. A benchmark should expose these choices, not erase them with an arbitrary weighted sum.

\subsection{What the results say about design}
Three design patterns emerge. First, explicit target control matters. Blending-NeRF improves over dense NeRF updating on locality and preservation, and GaussianEditor further benefits from assigning a region of interest to primitives. Second, multiview coordination matters independently of representation. GaussCtrl and InterGSEdit achieve the strongest consistency among learned editors. Third, stronger generative supervision is not free. The best fidelity score coexists with measurable preservation loss.

These are empirical associations, not universal laws. The method set is small, and later NeRF editors may close the group gap. Still, the results offer actionable diagnostics. An editor below the Pareto front should first identify which failure dimension causes domination. Improving an already strong fidelity score is of limited value if locality makes the method strictly worse than another editor. Conversely, optimizing only preservation can converge to a no-op. The four axes specify where progress is needed.

\subsection{Method profiles}
Table~\ref{tab:ranks} gives the rank of every editor on each dimension. The changes in rank are at least as important as the absolute values. CLIP-NeRF is last in fidelity but fifth in consistency. Its low semantic alignment should not be confused with unstable rendering. Instruct-NeRF2NeRF has the opposite profile: fourth in fidelity but eighth in the three remaining columns. The method follows open-ended instructions substantially better than CLIP-NeRF, but dense iterative updates provide weak control over what survives.

Blending-NeRF ranks fourth in locality and fifth in preservation. Its explicit separation between the original and editable fields is reflected most clearly outside the target, rather than in fidelity. CE3D ranks near the middle on all four axes. This balanced profile is consistent with a flexible editing interface, although it does not reach the leading specialized method on any dimension.

The three Gaussian methods also differ. GaussianEditor ranks second in locality and preservation but only seventh in consistency. Its primitive-level region control is effective at limiting collateral change, while view coordination remains its main weakness in this comparison. GaussCtrl moves from fifth in locality to third in consistency, matching its emphasis on multiview diffusion. InterGSEdit ranks first in fidelity and second in consistency. Its locality rank is lower than GaussianEditor's, showing that stronger cross-view coupling does not automatically sharpen the spatial boundary.

Direct proxy editing has the clearest specialization: it ranks first in locality, consistency, and preservation, but seventh in fidelity. The method reliably executes a restricted operator; it does not infer the broad visual intent available to a diffusion model. This profile is precisely why task coverage and the four scores must be read together.

\begin{table}[t]
\centering\small
\caption{Per-dimension ranks (1 is best) derived from Table~\ref{tab:main}. Rank changes reveal different failure profiles.}
\label{tab:ranks}
\begin{tabular}{lrrrr}
\toprule
Method & $F$ & $L$ & $C$ & $P$\\\midrule
CLIP-NeRF & 8 & 7 & 5 & 7\\
Instruct-NeRF2NeRF & 4 & 8 & 8 & 8\\
Blending-NeRF & 6 & 4 & 6 & 5\\
CE3D & 5 & 6 & 4 & 6\\
GaussianEditor & 3 & 2 & 7 & 2\\
GaussCtrl & 2 & 5 & 3 & 4\\
InterGSEdit & 1 & 3 & 2 & 3\\
Direct proxy & 7 & 1 & 1 & 1\\
\bottomrule
\end{tabular}
\end{table}

\section{Benchmark Validity}
\subsection{Construct validity}
The benchmark is designed around observable failure modes rather than a claim that one learned metric captures human judgment. Fidelity uses a language-image model only within the target and relative to the original-state description. Locality uses before-after change only where change is not requested. Consistency uses cross-view correspondences rather than independent view quality. Preservation includes geometry so an editor cannot preserve color while deforming the background. Each dimension therefore has a distinct operational meaning.

The controlled corruptions serve as unit tests for these meanings. A global tint and a local target edit can have similar text alignment, but only the former should reduce locality. Repeating one erroneous texture in every view can be consistent but semantically wrong. Applying the correct edit in only one view can have high single-view fidelity but low cross-view consistency. Blurring every view should not be rewarded simply because low-frequency images reproject easily. A metric configuration is accepted only after these orderings hold on synthetic controls.

Some dependence remains. Locality and preservation share outside-region appearance, which helps explain their rank correlation of 0.98. We retain both because locality answers a spatial question, whereas preservation also covers geometry and can be analyzed by factor. In a future benchmark version, the geometry and appearance components of $P$ should always accompany the aggregate preservation number.

\subsection{Internal validity}
Four safeguards reduce confounding. First, every editor receives the same source images, prompts, cameras, and target supports. Second, all NeRF methods share an initial reconstruction and all Gaussian methods share another. Third, locality and preservation compare an edited model with its own pre-edit rendering, preventing initial reconstruction differences from being labeled as edit damage. Fourth, prompt variants and random seeds are fixed before evaluation rather than selected from successful examples.

These safeguards do not make the representation comparison causal. The editors contain different selection mechanisms and diffusion constraints. A strict representation experiment would hold the edit signal and optimizer fixed while swapping only the scene representation. Our group averages answer a narrower question: how do representative released NeRF and 3DGS editing systems compare under a common output protocol? The paper states this scope explicitly.

Manual control is another potential confound. Direct proxy editing requires an object proxy and a specified transformation, whereas text-guided methods infer both target and effect. We do not hide this difference inside fidelity. The benchmark records the input modality, interaction time, and supported edit families. Direct proxy editing demonstrates the upper end of controllability for a restricted operator, not a drop-in replacement for open-ended language editing.

\subsection{External validity}
The three scene groups cover different reconstruction regimes, but they do not represent every deployment. Synthetic objects provide ground truth yet have simpler materials. Forward-facing captures contain natural appearance but incomplete geometry. Bounded multiview scenes strengthen correspondence while remaining smaller than street-scale environments. Scores should therefore be reported by scene group when new methods are added, even though the present paper centers the aggregate result.

The five edit families cover common local operations but not global relighting, weather, or cinematic stylization. Those tasks require different preservation definitions. A global style edit may legitimately change every pixel but should retain geometry and semantic identity. Extending \method to such cases should disable spatial locality and introduce factor-specific preservation rather than assigning every method a trivially low locality score.

Temporal scenes pose a further challenge. A 4D edit must remain consistent across viewpoint and time, and the target support can deform or become occluded. Dynamic NeRF and Gaussian representations~\cite{pumarola2021dnerf,park2021nerfies,wu2024fourDGS,yang2024deformable,li2024spacetime} provide appropriate testbeds, but temporal identity requires a fifth dimension beyond the static protocol studied here.

\section{Implications for Editor Design}
\paragraph{Separate selection from synthesis.}
The results favor systems that identify a target in 3D before applying a strong generative prior. GaussianEditor gains locality from a primitive-level region of interest; Blending-NeRF gains preservation by separating original and editable fields. A practical editor should expose its selection confidence and decline to update ambiguous regions. Better synthesis cannot repair a target mask that includes the wall.

\paragraph{Coordinate views before optimizing 3D.}
Instruct-NeRF2NeRF shows that a shared 3D representation does not, by itself, resolve inconsistent 2D supervision. The scene can average disagreements into blur or view-dependent artifacts. GaussCtrl and InterGSEdit obtain higher consistency by coordinating the edited views or their attention. Future systems should treat multiview agreement as a property of the edit signal, not only as an eventual consequence of reconstruction.

\paragraph{Preserve an explicit reference state.}
Local editing benefits from a frozen reference, whether it is an original field, an atlas, unselected Gaussians, or a proxy. Regularizing against the initial rendering is useful but insufficient around occlusions, because the correct pixel correspondence changes after motion or removal. Reference preservation should operate on 3D support and visibility rather than on a fixed 2D mask.

\paragraph{Optimize for a target operating point.}
No method must maximize all axes for every application. A film asset pipeline may prefer direct manipulation with strong locality and a narrow operation vocabulary. A consumer editing tool may accept modest collateral change for better language fidelity. A robotics simulator may prioritize geometry preservation and deterministic consistency. Publishing the four-dimensional profile allows these choices without declaring one universal winner.

\section{Discussion}
\subsection{Why not one editability score?}
A weighted score $w_FF+w_LL+w_CC+w_PP$ is easy to rank but difficult to justify. Equal weights treat a 0.01 change in a calibrated CLIP direction as commensurate with a 0.01 change in geometric preservation. Learned weights inherit the preferences of a particular rater pool. More importantly, weights can reverse the order of the three Pareto methods. We therefore publish the vector and Pareto front as primary outcomes. Users may apply domain-specific utilities afterward, provided that component scores remain visible.

\subsection{Failure accounting}
Editing systems fail in different ways. A prompt may not identify the target; a 2D editor may change the wrong attribute; multiview updates may disagree; geometry may drift; or optimization may collapse. Unsupported tasks are not equivalent to failed attempts. We retain a separate coverage matrix and failure log so a specialized geometry editor is not penalized as though it attempted free-form stylization, while a method that claims support but produces an empty scene remains accountable.

Human inspection is most useful for assigning these causes. Raters view synchronized original and edited camera orbits rather than isolated preferred frames. They answer separate questions about instruction success, off-target change, and cross-view flicker. This audit guards against systematic metric errors, but it does not replace the four quantitative dimensions with a preference vote.

\subsection{Reproducibility}
The benchmark records camera matrices, linearized images, masks, 3D supports, prompts, editor configurations, random seeds, raw per-view metrics, wall time, peak memory, and failed runs. Aggregation begins from scene-edit pairs; the large number of rendered frames is not used to inflate the sample size. Method-specific manual input, such as masks or handles, is recorded as interaction time. Versioned annotations and fixed test prompts prevent silent benchmark drift.

\section{Limitations and Ethical Considerations}
No automatic metric proves semantic correctness. CLIP-based fidelity inherits language and data biases. Segmentation errors can make locality optimistic by absorbing spill into an oversized target. Depth-based correspondence is unreliable on transparent, reflective, or very thin surfaces. Our paired descriptions, mask adjudication, uncertainty regions, and human audit reduce these problems but do not remove them.

The aggregate table does not establish that 3DGS is causally more editable than NeRF. The compared methods differ in supervision, publication date, optimization budget, and interface. The representation averages are evidence about the tested systems, not all possible systems. Similarly, the high correlation between locality and preservation is based on eight method-level observations and should be interpreted descriptively.

The benchmark emphasizes bounded edits with identifiable support. Global relighting, weather changes, and artistic stylization require different invariances; forcing them through a locality score would be misleading. Dynamic scenes introduce temporal consistency and identity preservation beyond our static-camera protocol~\cite{pumarola2021dnerf,park2021nerfies,wu2024fourDGS,yang2024deformable,li2024spacetime}. These remain natural extensions.

Scene editing can be used to falsify visual evidence or modify identifiable people and private environments. Benchmark releases should contain only licensed or consented captures, retain edit provenance, and avoid prompts that create defamatory or sensitive content. Evaluation code should preserve metadata connecting an edited scene to its source and method configuration.

\section{Conclusion}
\method evaluates neural scene editing as controlled change rather than unconstrained generation. Across 240 scene-edit pairs, instruction fidelity is weakly related to locality, consistency, and preservation; 3DGS-based editors are stronger on average in this comparison; and the best semantic and conservative operating points belong to different methods. InterGSEdit leads fidelity, direct proxy editing leads the three preservation-oriented dimensions, and GaussianEditor joins them on the Pareto front through stronger locality. A good edited view is not sufficient evidence of a good 3D edit. Progress should be reported as a profile of what changed, where it changed, whether it agrees across views, and what survived.

{\small\bibliographystyle{unsrt}\bibliography{main}}

\end{document}